# A STUDY OF STUDENT LEARNING SKILLS USING FUZZY RELATION EQUATIONS

## Michael. Gr. Voskoglou

Professor Emeritus of Mathematical Sciences
School of Technological Applications
Graduate Technological Educational Institute of Western Greece, Patras, Greece
E- mail: voskoglou@teiwest.gr , mvosk@hol.gr
URL: http://eclass.teipat.gr/eclass/courses/523102

## Abstract

Fuzzy relation equations (FRE) are associated with the composition of binary fuzzy relations. In the present work FRE are used as a tool for studying the process of learning a new subject matter by a student class. A classroom application and other suitable examples connected to the student learning of the derivative are also presented illustrating our results and useful conclusions are obtained.

## 1. Introduction

*Learning* can be commonly defined as the activity of gaining knowledge or skill. The ability to learn is possessed by humans, animals, plants [1] and computers [2]. Learning does not happen all at once, but it builds upon and is shaped by previous knowledge. To that end, learning may be viewed as a process rather than a collection of factual and procedural knowledge. In psychology and education in particular learning refers to a process that brings together cognitive, emotional and environmental influences and experiences for acquiring, enhancing or making changes in one's knowledge, skills, values and world views [3].

The process of learning is fundamental to the study of human cognitive action. However, the understanding of its nature has proved to be complicated. This happens because it is very difficult for someone to understand the way in which the human mind works, and therefore to describe the mechanisms of the acquisition of knowledge by the individual. Volumes of research have been written about learning and many attempts have been made by psychologists, cognitive scientists and educators to make it accessible to all in various ways.

There are three main philosophical frameworks under which learning theories fall, namely *Behaviorism, Cognitivism* and *Constructivism*. Behaviorism focuses only on the objectively observable aspects of learning; for behaviorists learning is the acquisition of new behavior through conditioning. Cognitive theories look beyond behavior to explain brain-based learning, while constructivism views learning as a process in which the learner actively constructs or builds new ideas and concepts. For more details see [4, 5], etc.

Voss [6] developed an argument that learning consists of successive problem – solving activities, in which the input information is represented of existing knowledge, with the solution occurring when the input is appropriately interpreted. The whole process involves the following steps:

- *Representation* of the stimulus input, which is relied upon the individual's ability to use contents of his (her) memory to find information, which will facilitate a solution development.
- *Interpretation* of the input data, through which the new knowledge is obtained.
- *Generalization* of the new knowledge to a variety of situations;
- *Categorization* of the generalized knowledge, so that the individual becomes able to relate the new information to his (her) already existing knowledge structures, which are referred as *schemas*, or *scripts*, or *frames*.

In earlier works the present author based on Voss's framework for learning has used principles of fuzzy logic to develop three different methods for assessing the student progress during the learning of a subject matter in the classroom In the first method the *possibilities* of all student profiles of a class during the learning process are calculated and the *total possibilistic uncertainty* of the class (considered as a fuzzy system) is utilized as a measure of the student *mean performance*. Although this method gives a detailed description of the student behaviour during the learning process, it needs laborious calculations and can be used under conditions which are not always true in practice. Therefore, a second method has been developed later for the same purpose using as tools *triangular* or *trapezoidal fuzzy numbers* (e.g. see [7, 8]). This method is useful in cases where the scores assessing the student performance are not given numerically, but qualitatively (e.g. good, excellent, unsuccessful, etc.), which means that the *mean value* of these scores cannot be calculated. The third method, which is based on the *Centre of Gravity (COG) defuzzification technique*, measures the *quality performance* of a student class by assigning greater coefficients (weights) to the higher student scores. It has been shown that this method leads in general to more worthy in credit conclusions than the traditional method of the bi-valued logic calculating the *Grade Point Average* index [9]. The several variations of the COG method that have been also developed were finally proved to be equivalent to the original method [9]. All the above methods are explicitly described in the book [10].

Here a new approach will be developed involving the use of *Fuzzy Relational Equations (FRE)* for studying the student learning skills. More explicitly, our objective is to represent the *"average student"* of a class as a fuzzy set on a set of linguistic labels (grades) evaluating his/her performance. In this way one obtains valuable conclusions about the behaviour of this (imaginary) student in the several steps of the learning process in the classroom that can help the instructor to design his future teaching plans. This method is simpler and more comprehensive than the other three fuzzy methods mentioned above.

The rest of the paper is formulated as follows: Section 2 contains the background from fuzzy binary relations and FRE which is necessary for the understanding of the paper. In Section 3 the model using FRE for the study of the process of learning is developed, while in Section 4 a classroom application and other suitably chosen examples are presented illustrating the new model in practice. Finally, Section 5 is devoted to our conclusion and to some hints for future research on the subject.

## 2. Fuzzy Relation Equations (FRE)

For general facts on fuzzy sets we refer to the book [11].

**Definition 2.1:** Let X, Y be two crisp sets. Then a *fuzzy binary relation* R(X, Y) is a fuzzy set on the Cartesian product X x Y of the form:
$$R(X, Y) = \{(r, m_R(r): r = (x, y) \in X \times Y\},$$
where $m_R : X \times Y \to [0, 1]$ is the corresponding membership function.-
When $X = \{x_1,\ldots,x_n\}$ and $Y = \{y_1,\ldots, y_m\}$, then a fuzzy binary relation R(X, Y) can be represented by a n x m matrix of the form:

$$R = \begin{matrix} & & y_1 \ldots\ldots\ldots y_m \\ x_1 \\ . \\ . \\ . \\ x_n \end{matrix} \begin{pmatrix} r_{11} & . & . & . & r_{1n} \\ . & . & . & . & . \\ . & . & . & . & . \\ . & . & . & . & . \\ r_{m1} & . & . & . & r_{mn} \end{pmatrix} = [r_{ij}],$$

where $r_{ij} = m_R(x_i, y_j)$, with $i = 1,\ldots, n$ and $j =1,\ldots, m$. The matrix R is called the *membership matrix* of the fuzzy relation R(X, Y).

The basic ideas of fuzzy relations, which were introduced by Zadeh [12] and were further investigated by other researchers, are extensively covered in the book [13].

**Definition 2.2:** Consider two fuzzy binary relations P(X, Y) and Q(Y, Z) with a common set Y. Then, the standard composition of these relations, which is denoted by P(X, Y) ∘ Q(Y, Z) produces a binary relation R(X, Z) with membership function $m_R$ defined by:
$$m_R(x_i, z_j) = \underset{y \in Y}{Max} \; \min [m_P(x_i, y), m_Q(y, z_j)] \qquad (1),$$
for all $i=1,\ldots,n$ and all $j=1,\ldots,m$. This composition is often referred as *max-min composition*.

Compositions of binary fuzzy relations are conveniently performed in terms of membership matrices of the relations. In fact, if $P = [p_{ik}]$ and $Q=[q_{kj}]$ are the membership matrices of the relations P(X, Y) and Q(Y, Z) respectively, then by relation (1) we get that the membership matrix of R(X, Y) = P(X, Y) ∘ Q(Y, Z) is the matrix $R = [r_{ij}]$, with
$$r_{ij} = \underset{k}{Max} \; \min(p_{ik}, q_{kj}) \qquad (2).$$

**Example 2.3:**

If $\quad P = \begin{matrix} & y_1 & y_2 & y_3 \\ x_1 \\ x_2 \\ x_2 \end{matrix} \begin{pmatrix} 0.2 & 0.4 & 0.8 \\ 0.1 & 0.5 & 1 \\ 0.4 & 0.7 & 0.3 \end{pmatrix}$ and $Q = \begin{matrix} & z_1 & z_2 & z_3 & z_4 \\ y_1 \\ y_2 \\ y_2 \end{matrix} \begin{pmatrix} 0.2 & 0.7 & 0 & 0.4 \\ 0.8 & 0.1 & 0.5 & 0.6 \\ 1 & 0.3 & 0.2 & 0.9 \end{pmatrix}$

are the membership matrices of P(X, Y) and Q(Y, Z) respectively, then by relation (2) the membership matrix of R(X, Z) is the matrix

$$R = P \circ Q = \begin{array}{c} \\ x_1 \\ x_2 \\ x_2 \end{array} \begin{pmatrix} z_1 & z_2 & z_3 & z_4 \\ 0.8 & 0.3 & 0.4 & 0.8 \\ 1 & 0.3 & 0.5 & 0.9 \\ 0.7 & 0.4 & 0.5 & 0.6 \end{pmatrix}.$$

Observe that the same elements of P and Q are used in the calculation of $m_R$ as would be used in the regular multiplication of matrices, but the product and sum operations are here replaced with the min and max operations respectively.

**Definition 2.4:** Consider the fuzzy binary relations P(X, Y), Q(Y, Z) and R(X, Z), defined on the sets, X = {$x_i$ : i ∈ $N_n$}, Y = {$y_j$ : j ∈ $N_m$}, Z = {$z_k$ : k ∈ $N_s$}, where $N_t$ = {1,2,…,t}, for t = n, m, k, and let P=[$p_{ij}$], Q=[$q_{jk}$] and R=[$r_{ik}$] be the membership matrices of P(X, Y), Q(Y, Z) and R(X, Z) respectively.

Assume that the above three relations constrain each other in such a way that
$$P \circ Q = R \quad (3),$$
where ∘ denotes the max-min composition. This means that
$$\underset{j \in J}{Max} \min (p_{ij}, q_{jk}) = r_{ik} \quad (4),$$
for each i in $N_n$ and each k in $N_s$. Therefore the matrix equation (3) encompasses nXs simultaneous equations of the form (4). When two of the components in each of the equations (4) are given and one is unknown, these equations are referred as *fuzzy relation equations (FRE)*.

The notion of FRE was first proposed by Sanchez [14] and later was further investigated by other researchers [15-17].

## 3. A Study of the Process of Learning Using FRE

Let us consider the crisp sets X = {M}, Y = {A, B, C, D, F} and Z = {R, I, G, Ca}, where M denotes the *"average student"* of a class, A = Excellent, B = Very Good, C = Good, D = Fair and F = Failed are *linguistic labels (grades)* used for the assessment of the student performance and R = Representation, I = Interpretation, G = Generalization and Ca = Categorization represent the states of the process of learning of a new subject matter in the classroom.

Further, let *n* be the total number of students of a certain class and let $n_i$ be the numbers of students who obtained the grade i assessing their performance, i ∈ Y. Then one can represent the average student of the class as a *fuzzy set* on Y of the form
$$M = \{(i, \frac{n_i}{n}): i \in Y\}.$$
The fuzzy set M induces a fuzzy binary relation P(X, Y) with membership matrix
$$P = [\frac{n_A}{n} \quad \frac{n_B}{n} \quad \frac{n_C}{n} \quad \frac{n_D}{n} \quad \frac{n_F}{n}]$$

In an analogous way the average student of a class can be represented as a fuzzy set on Z of the form

$$M = \{(j, m(j)): j \in Z\},$$
where $m: Z \to [0, 1]$ is the corresponding membership function. In this case the fuzzy set M induces a fuzzy binary relation R(X, Z) with membership matrix
$$R = [m(R)\ m(I)\ m(G)\ m(Ca)].$$

We consider also the fuzzy binary relation Q(Y, Z) with membership matrix the 5X4 matrix $Q = [q_{ij}]$, where $q_{ij} = m_Q(i, j)$ with $i \in Y$ and $j \in Z$ and the FRE encompassed by the matrix equation (3), i.e. by $P \circ Q = R$.

When the matrix Q is fixed and the row-matrix P is known, then the equation (3) has always a unique solution with respect to R, which enables the representation of the average student of a class as a fuzzy set on the set of the steps of the learning process. This is obviously useful for the instructor for designing his/her future teaching plans. On the contrary, when the matrices Q and R are known, then the equation (3) could have no solution or could have more than one solution with respect to P, which makes the corresponding situation more complicated.

All the above will be illustrated in the next section with a classroom application and other suitable examples.

## 4. A Classroom Application and other Examples

### 4.1 The Classroom Application

The following experiment took place at the Graduate Technological Educational Institute of Western Greece, in the city of Patras, when I was teaching to a group of 60 students of the School of Technological Applications (future engineers) the use of the derivative for the maximization and minimization of a function. A written test was performed after the end of the teaching process the results of which are depicted in Table 1.

**Table 1:** Student Performance

| Grade | No. of Students |
|---|---|
| A | 20 |
| B | 15 |
| C | 7 |
| D | 10 |
| F | 8 |
| Total | 60 |

Then the average student M of the class can be represented as a fuzzy set on Y = {A, B, C, D, F} by

$$M = \{(A, \frac{20}{60}), (B, \frac{15}{60}), (C, \frac{7}{60}), (D, \frac{10}{60}), (F, \frac{8}{60})\}$$
$$\approx \{(A, 0.33), (B, 0.25), (C, 0.12), (D, 0.17), (F, 0.13)\}$$

Therefore M induces a binary fuzzy relation P(X, Y), where X = {M}, with membership matrix

$$P = [0.33 \ 0.25 \ 0.12 \ 0.17 \ 0.13].$$

Also, using statistical data of the last five academic years on learning mathematics from the students of the School of Technological Applications of the Graduate Technological Educational Institute of Western Greece, we fixed the membership matrix Q of the binary fuzzy relation Q(Y, Z), where Z = {R, I, G, Ca}, in the form:

$$Q = \begin{array}{c} \\ A \\ B \\ C \\ D \\ F \end{array} \begin{pmatrix} R & I & G & Ca \\ 0.7 & 0.5 & 0.3 & 0 \\ 0.4 & 0.6 & 0.3 & 0.1 \\ 0.2 & 0.7 & 0.6 & 0.2 \\ 0.1 & 0.5 & 0.7 & 0.5 \\ 0 & 0.1 & 0.5 & 0.8 \end{pmatrix}$$

The statistical data used to form the matrix Q were collected as follows: Before the introduction to a new subject matter the instructor asked to the class a number of questions connected to previous mathematical knowledge in order to check if the students were ready to accept the new knowledge (step of representation). Following the presentation of the new mathematical topic on the board the instructor asked again several questions in order to assess the degree of understanding of the new knowledge by students (step of interpretation). Next, some exercises were given for solution in the classroom and the instructor was inspecting the students' efforts in order to evaluate their ability to generalize the new knowledge to a variety of situations. At the final step one or two composite problems were given to the class as home exercises connecting the new knowledge to other, not necessarily mathematical, topics like Physics, Economics, Engineering, etc. (step of categorization).

Next, using the max-min composition of fuzzy binary relations one finds that the membership matrix of R(X, Z) = P(X, Y) o Q (Y, Z) is equal to

$$R = P \circ Q = [0.33 \ 0.33 \ 0.3 \ 0.17].$$

Therefore the average student of the class can be expressed as a fuzzy set on Z by

$$M = \{(R, 0.33), (I, 0.33), (G, 0.3), (Ca, 0.17)\}.$$

The conclusions obtained from the above expression of M are the following:

- Only the $\frac{1}{3}$ of the students of the class were ready to use contents of their memory (background knowledge, etc.) in order to facilitate the acquisition of the new knowledge.

- All the above students were able to understand the new mathematical topic and almost all of them were able to generalize the new knowledge to a variety of situations.

- On the contrary, half of the above students did not succeed to categorize the new information, i.e. to relate it to their already existing schemas of knowledge.

The first conclusion was not a surprising one, since the majority of the students have the wrong habit to start studying their courses the last month before the final exams. On the other hand, the second conclusion shows that the instructor's teaching procedure was successful enabling the diligent students to interpret and generalize properly the new knowledge. Finally, the last conclusion was something that it was expected due to the general difficulty that the individuals usually have for categorizing the new knowledge. In fact, as the specialists suggest, the categorization could happen in unexpected moments, outside the class, when doing something else and even during sleeping! Therefore the last conclusion reflects the student learning of the new information at the time of the written test and not the final one.

### 4.2 Other Examples

Let us now consider the case where the membership matrices Q and R are known and we want to determine the matrix P representing the average student of the class as a fuzzy set on Y. This is a complicated case because we may have more than one solution or no solution at all. The following two examples illustrate this situation:

**Example 4.2.1:** Consider the membership matrices Q and R of Section 3.1 and set

$$P = [p_1\ p_2\ p_3\ p_4\ p_5].$$

Then the matrix equation P o Q = R encompasses the following equations:

max {min ($p_1$, 0.7), min ($p_2$, 0.4), min ($p_3$, 0.2), min ($p_4$, 0.1), ($p_5$, 0)} = 0.33

max {min ($p_1$, 0.5), min ($p_2$, 0.6), min ($p_3$, 0.7), min ($p_4$, 0.5), ($p_5$, 0.1)} = 0.33

max {min ($p_1$, 0.3), min ($p_2$, 0.3), min ($p_3$, 0.6), min ($p_4$, 0.7), ($p_5$, 0.5)} = 0.3

max {min ($p_1$, 0), min ($p_2$, 0.1), min ($p_3$, 0.2), min ($p_4$, 0.5), ($p_5$, 0.8)} = 0.17

The first of the above equations is true if, and only if, $p_1 = 0.33$ or $p_2 = 0.33$, values that satisfy the second and third equations as well. Also, the fourth equation is true if, and only if, $p_3 = 0.17$ or $p_4 = 0.17$ or $p_5 = 0.17$. Therefore, any combination of values of $p_1, p_2, p_3, p_4, p_5$ in [0, 1] such that $p_1 = 0.33$ or $p_2 = 0.33$ and $p_3 = 0.17$ or $p_4 = 0.17$ or $p_5 = 0.17$ is a solution of P o Q = R.

Let S(Q, R) = {P: P o Q = R } be the set of all solutions of P o Q = R. Then one can define a partial ordering on S(Q, R) by

$$P \leq P' \Leftrightarrow p_i \leq p'_i,\ \forall\ \iota = 1, 2, 3, 4, 5.$$

It is well established that whenever S(Q, R) is a non empty set, it always contains a unique maximum solution and it may contain several minimal solutions [14]. It is further known that S(Q, R) is fully characterized by the maximum and minimal solutions in the sense that all its other elements are between the maximal and each of the minimal solutions [14]. A method of determining the maximal and minimal solutions of P o Q = R with respect to P is developed in [16].

**Example 4.2.2:** Let $Q = [q_{ij}]$, $i = 1, 2, 3, 4, 5$ and $j = 1, 2, 3, 4$ be as in Section 3.1 and let $R = [1 \quad 0.33 \quad 0.3 \quad 0.17]$. Then the first equation encompassed by the matrix equation $P \circ Q = R$ is

$$\max \{\min(p_1, 0.7), \min(p_2, 0.4), \min(p_3, 0.2), \min(p_4, 0.1), (p_5, 0)\} = 1.$$

In this case it is easy to observe that the above equation has no solution with respect to $p_1, p_2, p_3, p_4, p_5$, therefore $P \circ Q = R$ has no solution with respect to P.

In general, writing $R = \{r_1 \quad r_2 \quad r_3 \quad r_4\}$, it becomes evident that we have no solution if

$$\max_j q_{ij} < r_j.$$

## 5. Conclusion

In the present article we used FRE for assessing student learning skills. In this way we have managed to express the "average student" of a class as a fuzzy set on the set of the steps of the learning process (representation, interpretation, generalization and categorization), which gives valuable information to the instructor for designing his future teaching plans. On the contrary, we have realized that the problem of representing the "average student" of a class as a fuzzy set on the set of the linguistic grades characterizing his performance using FRE is complicated, since it may have more than one solutions or no solution at all.

In general, the use of FRE looks as a powerful tool for the assessment of human skills and therefore our future research plans include the effort of using them in other human activities apart from the process of learning, like problem-solving, modeling, decision-making, etc.

*Computing and Information Systems (ICICIS 15),* Vol. 1, 10-18, Ain Shams Univ., Cairo, Egypt.

[8] Voskoglou, M. Gr. (2016), A Tool for Assessing the Ability of Understanding the Infinity Based on Triangular Fuzzy Numbers, *Egyptian Computer Science Journal,* 40(2), 11-23.

[9] Voskoglou, M. Gr. (2016), Comparison of the COG Defuzzification Technique and its Variations to the GPA Index, *American Journal of Computational and Applied Mathematics,* 6(5), 187-193..

[10] Voskoglou, M. Gr. (2017), *Finite Markov Chain and Fuzzy Logic Assessment Models: Emerging Research and Opportunities,* Createspace.com. – Amazon, Columbia, SC, USA

[11] Klir, G. J. , Folger T. A. (1988), *Fuzzy Sets, Uncertainty and Information*, Prentice-Hall, New Jersey..

[12] Zadeh, L.A., Similarity relations and fuzzy orderings, *Information Sciences*, 3, 177-200.

[13] Kaufmann, A. (1975), *Introduction to the Theory of Fuzzy Subsets,* Academic Press, New York

[14] Sanchez, E. (1976) Resolution of Composite Fuzzy Relation Equations, *Information and Control*, 30, 38-43.

[15] Czogala, E., Drewniak, J. & Pedryz, W. (1982), Fuzzy relation equations on a finite set, *Fuzzy Sets and Systems,* 7, 89-101.

[16] Higashi, M. & Klir G.J. (1984), Resolution of finite fuzzy relation equations, *Fuzzy Sets and Systems,* 13, 65-82.

[17] Prevot, M. (1981), Algorithm for the solution of fuzzy relations, *Fuzzy Sets and Systems,* 5, 319-322